\documentclass[runningheads]{llncs}
\usepackage[T1]{fontenc}
\usepackage{graphicx}
\usepackage{graphicx}
\usepackage{comment}
\usepackage{amsmath,amssymb}
\usepackage{color}
\usepackage{url}
\usepackage{hyperref}
\usepackage{multirow}
\usepackage{booktabs}
\usepackage{bm}
\usepackage{siunitx}
\usepackage{tablefootnote}

\usepackage[dvipsnames]{xcolor}
\definecolor{domainred}{RGB}{170, 0, 0}
\definecolor{modalityblue}{RGB}{42,127,255}

\usepackage{biblatex}
\graphicspath{{img}}

\usepackage[acronym]{glossaries}

\newacronym{hsi}{HSI}{hyperspectral imaging}
\newacronym{hs}{HS}{hyperspectral}
\newacronym{cnn}{CNN}{Convolutional Neural Network}
\newacronym{ml}{ML}{machine learning}
\newacronym{bn}{BN}{batch normalization}
\newacronym{dr}{DR}{dimensionality reduction}
\newacronym{cw}{CW}{class weighting}
\newacronym{do}{DO}{dropout}
\newacronym{da}{DA}{data augmentation}
\newacronym{wd}{WD}{weight decay}
\newacronym{usm}{USM}{unsharp masking}
\newacronym{pca}{PCA}{principal component analysis}
\newacronym{pc}{PC}{principal component}
\newacronym{hcv}{HCV2}{HyperspectralCity V2.0}
\newacronym{ins}{INS}{inverse number of samples}
\newacronym{isns}{ISNS}{inverse square root of number of samples}
\newacronym{sgd}{SGD}{Stochastic Gradient Descent}
\newacronym{iou}{IoU}{intersection over union}
\newacronym{dl3}{DL3+}{DeeplabV3+}
\newacronym{prgb}{pRGB}{pseudo-RGB}
\newacronym{aspp}{ASPP}{atrous spatial pyramid pooling}
\newacronym{fwhm}{FWHM}{full-width at half-maximum}
\newacronym{mae}{MAE}{masked auto-encoder}

\newcommand{\hsidrive}{HSI-Drive}
\newcommand{\hyko}{HyKo2}

\newcommand{\fone}{\textnormal{F}_1}

\newcommand{\eg}{e.\,g.}
\newcommand{\ie}{i.\,e.}

\newacronym{oa}{OA}{overall accuracy}
\newacronym{aa}{AA}{average accuracy}
\newacronym{miou}{mIoU}{mean intersection over union}

\newcommand{\unet}{U-Net}
\newcommand{\runet}{RU-Net}
\newcommand{\hsl}{HyperSL}
\newcommand{\hslrunet}{HyperSL-RU-Net}
\newcommand{\hsiadapter}{HSI-Adapter}
\newcommand{\justoliunet}{1D-Justo-LiuNet}

\newcommand{\hdcmr}{HDC-MiniROCKET}
\newcommand{\mr}{MiniROCKET}
\newcommand{\dino}{DINOv2}
\newacronym{mlp}{MLP}{Multi-Layer Perceptron}

\begin{document}
\title{Cross-Domain Transfer of Hyperspectral Foundation Models}
%
%
\author{Nick Theisen\orcidID{0000-0002-7758-0379} \and
Peer Neubert\orcidID{ 0000-0002-7312-9935}}
\authorrunning{N. Theisen et al.}
%
\institute{
Intelligent Autonomous Systems, University of Koblenz, Germany\\
\email{\{nicktheisen,neubert\}@uni-koblenz.de}}
\maketitle              
\begin{abstract}
\Gls*{hsi} semantic segmentation typically relies on in-domain training, but limited data availability often restricts model performance in real-world applications. Current approaches to leverage foundation models in proximal sensing use cross-modality techniques, bridging RGB and \gls*{hsi} to exploit vision foundation models. However, these methods either discard spectral information or introduce architectural complexity.
We propose cross-domain transfer as an alternative, reusing \gls*{hsi} foundation models -- originally trained in remote sensing -- for proximal sensing applications. By eliminating the need to bridge modality gaps, our approach preserves spectral information while maintaining a simple architecture.
Using the HS3-Bench benchmark, we systematically evaluate and compare conventional in-domain, in-modality training, cross-modality transfer and cross-domain transfer strategies. Our results demonstrate that cross-domain transfer achieves large performance improvements over in-domain, in-modality training, reduces the performance gap to cross-modality approaches and maintains strong performance in limited data settings. Thus, this work advances more effective \gls*{hsi} semantic segmentation in diverse applications.

\end{abstract}

\graphicspath{{img/}}
\centerline{\footnotesize This paper was accepted for publication at International Conference of}
\centerline{\footnotesize Pattern Recognition (ICPR) 2026.}
\section{Introduction}

\begin{figure}
    \centering
    \includegraphics[width=\linewidth]{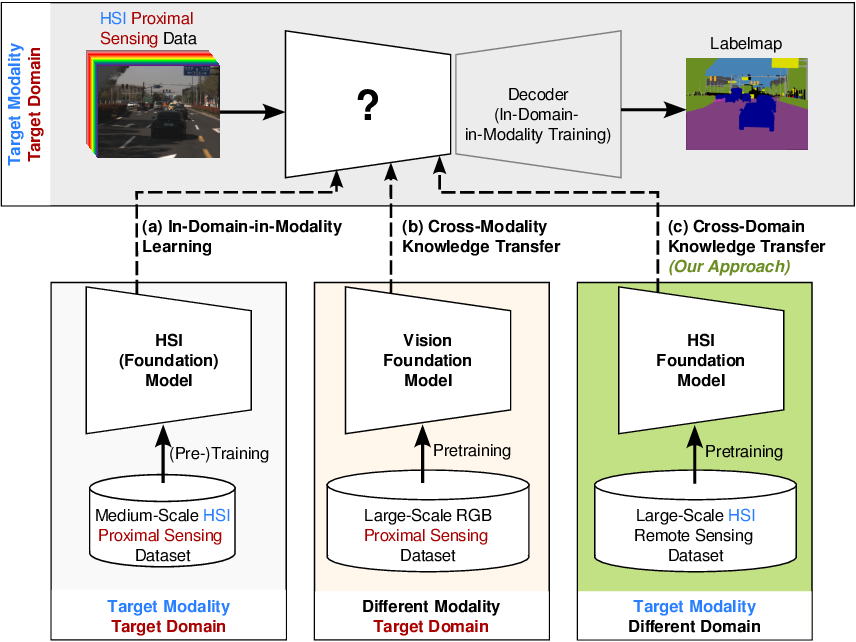}
    \caption{The typical approach for \gls*{hsi} semantic segmentation uses in-domain-in-modality training \textbf{(a)}, \ie\ training models on data from the target domain (\textcolor{domainred}{red}) and target modality (\textcolor{modalityblue}{blue}), but many applications face limited training data availability. An established method to address this problem is cross-modality knowledge transfer \textbf{(b)}, bridging RGB and \gls*{hsi} to exploit vision foundation models. These methods either sacrifice spectral information or increase architectural complexity. As an alternative, we introduce cross-domain knowledge transfer \textbf{(c)}, \ie\ exploiting \gls*{hsi} foundation models from a different domain -- eliminating the need to bridge modality gaps. We demonstrate the effectiveness of cross-domain approaches by systematic evaluation of all three strategies on the HS3-Bench benchmark.}
    \label{fig:visual-abstract}
\end{figure}

\gls*{hsi} systems offer perception capabilities beyond those of conventional RGB cameras. By sampling the electromagnetic spectrum across hundreds of narrow spectral bands -- including ranges invisible to the human eye, such as near-infrared and ultraviolet -- \gls*{hsi} enables detailed inference of material properties and surface textures. This makes \gls*{hsi} a powerful tool for tasks like semantic segmentation. However, leveraging hyperspectral data presents significant challenges.

Deep learning models currently define the state of the art, typically relying on in-domain-in-modality training (see \figurename~\ref{fig:visual-abstract} (a)), \ie\ training models on data similar to the target application domain. This approach demands large amounts of labeled training data. Yet, aquiring and annotating \gls*{hsi} data is costly and complex. Unlike RGB sensors, \gls*{hsi} systems remain expensive, require specialized expertise for calibration and processing and need domain-specific knowledge for accurate labeling. Sensor variability, \ie\ differences in spectral range, band count, central wavelengths and bandwidth, further complicates model transfer across devices. Thus, developing models that require less training data or can reuse knowledge, is critical for expanding the application of \gls*{hsi} to novel or niche domains.

A common alternative or extension to in-domain-in-modality training is cross-modality knowledge transfer (\figurename~\ref{fig:visual-abstract} (b)), leveraging vision foundation models, by bridging the gap between RGB and \gls*{hsi} data. While these models either sacrifice spectral information or increase architecture complexity, they have proven highly effective for semantic segmentation in driving scenes \cite{Theisen2024HBS,Hurtado2025HAS}. Surprisingly, models trained on \gls*{prgb} projections often outperform those using full spectral information, likely due to the maturity and robustness of RGB-based vision models \cite{Theisen2024HBS}.

Recent advances have introduced hyperspectral foundation models \cite{Hong2024SSR,Kong2025HSF,Wang2025HHI,Li2025HCT}, trained in self-supervised manner on large datasets. These models adress the challenge of limited training data by providing pretrained feature extractors for downstream tasks such as classification or change detection. However, these existing models are trained on remote sensing data, a domain where \gls*{hsi} is well-established due to decades of space- and airborne campaigns (\eg\ EO-1 or Gaofeng-5). In contrast, terrestrial applications -- \eg\ autonomous driving, agriculture, industrial inspection -- lack comparable public datasets, limiting the development of similar foundation models. To our knowledge \cite{Laprade2025GPS} is the only description of a hyperspectral foundational model with a focus on proximal sensing, which is unfortunately not publicly available, yet.

We investigate cross-domain knowledge transfer for \gls*{hsi} semantic segmentation (\figurename~\ref{fig:visual-abstract} (c)). This strategy reuses \gls*{hsi} foundation models trained in the remote sensing domain for proximal sensing applications, eliminating the need to bridge modality gaps. By preserving spectral information and maintaining architectural simplicity, it offers a promising alternative to existing methods.

The differences between remote sensing and terrestrial \gls*{hsi} data are substantial. Domain-specific challenges are introduced through atmospheric distortion, irregular object shapes and varying illumination conditions (\eg\ artificial vs. natural light). It remains unclear whether \gls*{hsi} foundation models trained on remote sensing data can generalize effectively to terrestrial applications. 

This raises the key question, which strategy is more effective for \gls*{hsi} semantic segmentation: (1) in-domain-in-modality training on limited datasets, (2) transferring hyperspectral foundation models across domains, or (3) exploiting well-established vision models?

We address this question in two steps:
\begin{enumerate}
    \item \textit{How can \gls*{hsi} semantic segmentation approaches be systematically organized based on the origin of their encoded knowledge?} We propose a taxonomy, that distinguishes between in-domain-in-modality training, cross-modality knowledge transfer and cross-domain knowledge transfer. It highlights cross-domain knowledge transfer as an approach that remains unexplored for this problem. 
    \item \textit{Does cross-domain knowledge transfer improve performance in \gls*{hsi} semantic segmentation?} Through systematic comparison using the HS3-Bench benchmark, we demonstrate that cross-domain knowledge improves model performance over in-domain-in-modality training and also narrows the performance gap to cross-modality approaches, which benefit from the rich RGB-data availability. Our model\footnote{The source code of our work is available under \url{https://github.com/nickstheisen/cross-domain-hsi}}, \hslrunet , which combines our previous model \runet\ \cite{Theisen2024HBS} with the \gls*{hsi} remote sensing foundational model \hsl , achieves state-of-the-art results among \gls*{hsi}-only methods and matches the performance of cross-modality approaches that do not rely on knowledge from RGB vision foundation models.
\end{enumerate}

The paper is structured as follows: Sec.~\ref{sec:related-work} summarized related work in \gls*{hsi} foundation models, Sec.~\ref{sec:aproaches} introduces the \gls*{hsi} semantic segmentation taxonomy and describes the cross-domain models, Sec.~\ref{sec:experiments} covers experimental results and Sec.~\ref{sec:conclusion} summarizes our findings.
\section{Related Work} \label{sec:related-work}

Recently, several hyperspectral foundational models have been proposed, primarily based on the Transformer architecture and focused on the remote sensing domain \cite{Hong2024SSR,Wang2025HHI,Li2025HCT,Kong2025HSF}. To date, the only proximal sensing foundation model, proposed by \citeauthor{Laprade2025GPS} \cite{Laprade2025GPS}, was trained on a collection of small-scale datasets but is not yet available to the public.

The first multi-spectral foundation model, SatMAE, was introduced by \citeauthor{Cong2022Spt} \cite{Cong2022Spt}. For pretraining it uses self-supervised learning via masked reconstruction on unlabeled satellite data. SpectralGPT \cite{Hong2024SSR} later addressed some of SatMAE's limitations, such as inconsistencies in spectral continuity caused by adjacent band grouping. \citeauthor{Ligan2025PFM} \cite{Ligan2025PFM} presented a framework for SpectralGPT to improve efficiency during finetuning.

HyperSIGMA \cite{Wang2025HHI} employs a two-stream architecture: one stream extracts spectral features the other spatial features, which are then combined through late fusion in an attention module. It accommodates \gls*{hsi} data with varying number of channels via random channel cropping. The authors also compiled the HyperGlobal-450K dataset, a large-scale collection of \gls*{hsi} remote sensing images. The SpectralEarth datast \cite{Braham2024STH} further expanded training data availability. HyperFree \cite{Li2025HCT} presents a tuning-free foundational model, dynamically building an embedding layer from task-specific prompts, demonstrating effectiveness across various downstream tasks in remote sensing.

The \hsl\ foundation model \cite{Kong2025HSF} introduces the spectral tokenizer module paired with a Transformer-based encoder-decoder. The tokenizer extracts spectral tokens from the input signal and combines them with a positional encoding based on each channel's spectral wavelength. This design allows the model to serve as a backbone for downstream tasks involving \gls*{hsi} images with arbitrary channels and spectral ranges, without requiring finetuning. Thus, we adopt \hsl\ in our work.
\section{Cross Domain Models as an Alternative Approach for HSI Semantic Segmentation} \label{sec:aproaches}
\subsection{HSI Semantic Segmentation Taxonomy} \label{sec:taxonomy}

In a typical \gls*{hsi} semantic segmentation scenario, a model processes data from a target modality (\gls*{hsi}) to solve a specific task within a target domain.

\begin{itemize}
    \item \textit{Task} refers to a specific problem defined by an input and a desired output, such as semantic segmentation or anomaly detection. In this sense, spectral classification (inferring a class from a predefined set of classes for a spectral vector) can be used to solve the same task as \gls*{hsi} semantic segmentation (inferring a class from a predefined set of classes for each pixel in an \gls*{hsi} image) by applying spectral classification pixel wise. However, semantic segmentation can exploit spatial context, while spectral classification relies solely on spectral information.

    \item \textit{Domain} refers to the application field from which the data originates, \eg\ remote sensing or proximal sensing. The domain shapes the data's characteristics and distribution.

    \item \textit{Modality} refers to the sensor class used for data acquisition, \eg\ RGB or \gls*{hsi}. The modality defines the data's structure and affects its distribution. We use this term broadly and do not distinguish between \gls*{hsi} sensors covering different spectral ranges.
\end{itemize}

Our proposed taxonomy, categorizes \gls*{hsi} semantic segmentation approaches into three groups based on the origin of their encoded knowledge as shown in \figurename~\ref{fig:visual-abstract}.

\begin{enumerate}
    \item \textit{In-Domain-In-Modality Training} (\figurename~\ref{fig:visual-abstract} (a)) is the canonical approach for semantic segmentation. It uses training data from the same domain and modality as the target application. While this avoids the need for modality or domain alignment, it requires substantial labeled data in the target domain and modality, which is often scarce.
    
    \item \textit{Cross-Modality Knowledge Transfer} (\figurename~\ref{fig:visual-abstract} (b)) leverages knowledge from a different modality. For \gls*{hsi} semantic segmentation it bridges the gap between RGB and \gls*{hsi} to exploit vision foundation models. The modality gap can be bridged by reducing dimensionality to three channels, \eg\ \cite{Winkens2019DDR,Theisen2024HBS}, resulting in information loss, or using specific \hsiadapter\ modules, \eg \cite{Hurtado2025HAS}, increasing architectural complexity. The success of cross-modality knowledge transfer depends on aligning vision foundation models with the target modality, which is challenging when the target modality's spectral range differs from the visual spectrum. 
    
    \item \textit{Cross-Domain Knowledge Transfer} (\figurename~\ref{fig:visual-abstract} (c)) leverages knowledge from a source domain in a target domain. In our case, this involves transferring \gls*{hsi} foundation models from the remote sensing domain to driving scenes. These models, designed for \gls*{hsi} data, preserve spectral information and can be directly used as backbones, maintaining a simpler architecture. The success of cross-domain knowledge depends on aligning \gls*{hsi} foundation models from source domain with data from target domain, which can be difficult when the data distributions of source and target domains differ significantly.
\end{enumerate}

\subsection{Cross-Domain Models} \label{sec:models}

To demonstrate and evaluate the potential benefit from cross-domain knowledge transfer, we implement this idea in two novel models. Both use \hsl\ \cite{Kong2025HSF} as a backbone to transfer knowledge from the \gls*{hsi} remote sensing domain to solve a semantic segmentation task in the \gls*{hsi} proximal sensing domain. 

\begin{figure}[h]
    \centering
    \includegraphics[width=0.9\linewidth]{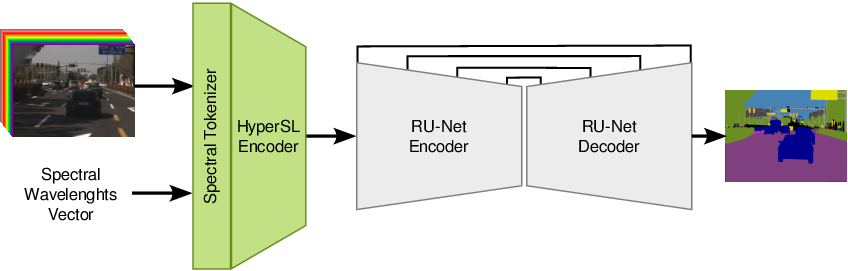}
    \caption{\hslrunet\ consists of a \hsl\ \cite{Kong2025HSF} backbone and an \runet\ \cite{Theisen2024HBS} encoder-decoder module for semantic segmentation.}
    \label{fig:hslrunet}
\end{figure}

The first model, \hslrunet\ (\figurename~\ref{fig:hslrunet}), is a semantic segmentation model. For feature extraction it uses the pretrained spectral tokenizer and encoder module from the \hsl\ foundation model and for classification our previous model \runet\ \cite{Theisen2024HBS}, a regularized version of \unet. We froze the parameters of the \hsl\ backbone and finetuned only the neck and classification modules on in-domain-in-modality data. This allows us to compute \hsl\ features once, improving efficiency by avoiding redundant calculations in each training epoch. The spectral tokenizer uses each channels spectral wavelength for positional encoding. Thus, we provide a wavelengths vector as additional input, consisting of the central wavelengths for each band of the target domain data.

We configured the \hsl\ encoder to process individual pixels, instead of patches, focusing on spectral information and ensuring applicability in spectral-only scenarios. \hslrunet\ additionally exploits spatial information through convolutional blocks in the \runet\ module.

\begin{figure}[h]
    \centering
    \includegraphics[width=0.9\linewidth]{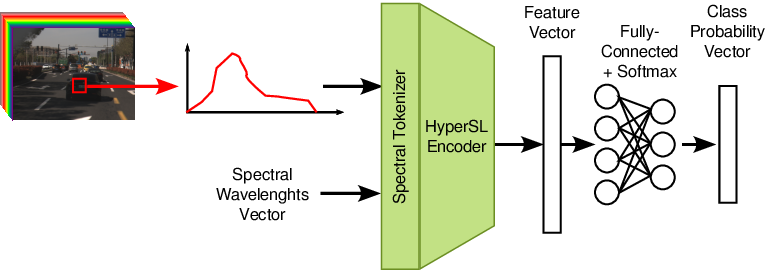}
    \caption{\hsl-FC consists of a \hsl\ \cite{Kong2025HSF} backbone and and a fully connected layer followed by softmax activation for spectral classification.}
    \label{fig:hsl-fc}
\end{figure}

The second model, \hsl-FC (\figurename~\ref{fig:hsl-fc}), is a spectral classification model. In contrast to \hslrunet , \hsl-FC takes individual pixel spectra and a corresponding wavelength vector as input and uses the same \hsl\ encoder as \hslrunet\ to compute a feature vector. The encoded feature vector is then given to a classification module, consisting of a fully connected layer with softmax activation. \hsl-FC was inspired by the spectral classification model, \hsl-CLF, proposed by the authors of \hsl\ \cite{Kong2025HSF}. We replace \hsl-CLF's more sophisticated classification module with a fully connected layer for better comparability with existing spectral classification models, that we use as a baseline in our experiments.

\section{Experiments \& Results} \label{sec:experiments}

In this section we first evaluate the effectiveness of cross-domain knowledge transfer for \gls*{hsi} semantic segmentation and spectral classification in Sec. \ref{subsec:domain-transfer-ss}. We then compare cross-domain to cross-modality knowledge transfer in Sec. \ref{subsec:rgb-models} and finally investigate the data-efficiency of our cross-domain models in Sec. \ref{subsec:data-efficiency}.

\subsection{Experimental Setup} \label{subsec:experimentalsetup}

We followed the HS3-Bench protocol for hyperspectral semantic segmentation in driving scenes \cite{Theisen2024HBS}. The benchmark includes three datasets: \gls*{hcv} \cite{Li2022HCV}, \hyko\ \cite{Winkens2017Hyko} and \hsidrive\ \cite{Basterretxea2021HSIDrive}. Dataset characteristics are summarized in \tablename~\ref{tab:used_datasets}. During training, we used only the training split, reserving validation split for hyperparameter tuning and early stopping. All results are based on the held-out test set, with fixed dataset splits as provided in \cite{Theisen2024HBS}.

\begin{table}[h]
    \footnotesize

    \caption{Overview of the \gls*{hsi} datasets used in HS3-Bench}
    \centering
    \resizebox{0.7\columnwidth}{!}{%
    \begin{tabular}{c|ccc}
        Name & \hyko & \gls*{hcv} & \hsidrive \\\hline
        Image size & $254 \times 510$ & $1400 \times 1800$ & $409 \times 216$ \\
        Bands & $15$ & $128$ & $25$ \\
        Range (nm) & $470$-$630$ & $450$-$950$ & $600$-$975$ \\
        Images & $371$ & $1330$ & $752$ \\
        Classes & $10$ & $19$ & $9$ \\
        Train/Test/Val-split ($\%$) & $50$/$20$/$30$ & $72$/$8$/$20$ & $60$/$20$/$20$\\
    \end{tabular}
    }
        \label{tab:used_datasets}
\end{table}

\citeauthor{Kong2025HSF} \cite{Kong2025HSF} provide three sets of pretrained weights for \hsl. Preliminary tests showed no significant performance differences among pretrained weights, so we used \textit{5\_base\_mask95\_checkpoint.pt} for all experiments. 

All experiments used a fixed random seed of 42. The semantic segmentation hyperparameters followed HS3-Bench, except for setting the maximum epochs for \hyko\ to 300 and the batch size for \hsidrive\ to 16 to reduce training time. For spectral classification, we set maximum epochs to 100 for all datasets and and used learning rates of 1e-1, 3e-4, 3e-4, 1e-4 and 1e-4 for \justoliunet\ \cite{Justo2025SSS}, \mr\ \cite{Dempster2021MVF}, \hdcmr\ \cite{Schlegel2022HET}, \hsl-FC and \hsl , respectively. We used 1000 features for \mr-based models and set the scale parameter to 5 for \hdcmr . Training was conducted on a single NVIDIA H100 GPU (80 GB VRAM).

For evaluation we use \gls*{oa}, \gls*{aa}, F1-Score ($\fone$) and \gls*{miou} metrics, which are equivalent to $\textnormal{Acc}_\mu$, $\textnormal{Acc}_M$, $\textnormal{F1}_M$ and $\textnormal{J}_M$ in \cite{Theisen2024HBS} as well as aAcc, mAcc and mIoU in \cite{Hurtado2025HAS}.

\subsection{Effectiveness of Cross-Domain Transfer} \label{subsec:domain-transfer-ss}
We first evaluated the effectiveness of cross-domain knowledge transfer for \gls*{hsi} semantic segmentation and for spectral classification. While the former leverages spatial context, the latter relies solely on spectral information.

\begin{table}[h!]
    \centering
    \caption{Benchmark scores (\%) on the HS3-Bench \cite{Theisen2024HBS} test data for \gls{hsi}-only semantic segmentation models, \ie\ hyperspectral data as input and hyperspectral backbone models. Best results are bold.}
    \hspace*{-0.7cm}

    \begin{tabular}{llcr|cccc}
                    \toprule
                        
                       &&&& \multicolumn{4}{c}{Testing}  \\\cmidrule(lr){5-8}
                       Dataset & Approach & Backbone & Data & \acrshort{oa} & \acrshort{aa} & $\fone$ & \acrshort{miou}
                       \\\midrule
                       \gls*{hcv} & \unet & -- & \gls*{hsi} & 85.25 & 48.62 & 48.18 & 37.73\\
                                 & \runet\  & -- & \gls*{hsi} & 87.63 & 54.14 & 53.26 & 42.23\\
                                 & \runet\ & \hsl\ & \gls*{hsi} & \textbf{88.17} & \textbf{57.53} & \textbf{55.12} & \textbf{43.92} \\
                                 \midrule
                       \hyko & \unet & -- & \gls*{hsi} & 85.36 & 68.15 & 68.55 & 57.39\\
                                 & \runet\  & -- & \gls*{hsi} & 86.72 & 68.79 & 69.19 & 58.64\\
                                 & \runet\ & \hsl\ & \gls*{hsi} & \textbf{88.24} &  \textbf{71.21} & \textbf{71.78} & \textbf{61.54} \\
                                \midrule
                       \hsidrive & \unet & -- & \gls*{hsi} & 94.95 & 74.74 & 76.08 & 64.95 \\
                                 & \runet\ & -- & \gls*{hsi} & 96.08 & 79.82 & 82.34 & 72.18\\
                                 & \runet\ & \hsl\ & \gls*{hsi} & \textbf{96.78} & \textbf{84.14} & \textbf{85.60} & \textbf{76.33}  \\
                                 \midrule
                       \textbf{Average}     & \unet & -- & \gls*{hsi} & 88.52 & 63.84 & 64.27 & 53.36\\
                       \textbf{Perf.} & \runet\ & -- & \gls*{hsi} & 90.14 & 67.58 & 68.26 & 57.68 \\
                                 & \runet\ & \hsl\ & \gls*{hsi} & \textbf{91.06} & \textbf{70.89} & \textbf{70.83} & \textbf{60.60} \\

                       \midrule
                       \textbf{Worst-Case}  & \unet  & -- & \gls*{hsi} & 82.25 & 48.63 & 48.18 & 37.73\\
                       \textbf{Perf.} & \runet  & -- & \gls*{hsi} & 86.72 & 54.14 & 53.26 & 42.23 \\
                                    & \runet\ & \hsl\  & \gls*{hsi} & \textbf{88.17} & \textbf{57.53} & \textbf{55.12} & \textbf{43.92} \\
                       \midrule
                       \bottomrule
    \end{tabular}
    \label{tab:hs3-hyper-ss-comparison}
\end{table}

To asses the benefit of cross-domain knowledge transfer for \gls*{hsi} semantic segmentation, we compare \hslrunet\ to two baseline models: A vanilla \unet\ \cite{Ronneberger2015UCN} and a regularized \unet\ (\runet ) \cite{Theisen2024HBS}, both trained from scratch on \gls*{hsi} data. 

The results are shown in \tablename~\ref{tab:hs3-hyper-ss-comparison}. \hslrunet\ outperforms both baselines across all datasets, with an average improvement of around 3\% in \gls*{miou}. Gains in class-averaged metrics (\gls*{aa}, $\fone$, \gls*{miou}) were more pronounced than in sample-averaged metric \gls*{oa}, indicating \hsl 's effectiveness for minority class prediction. These results demonstrate the effectiveness of cross-domain knowledge transfer and show that \hslrunet\ achieves state-of-the-art performance for \gls*{hsi}-only models.\\

To assess the benefit of cross-domain knowledge transfer for spectral-only approaches, \ie\ spectral classification, we compare two models using a \hsl\ backbone (\hsl-FC, \hsl-CLF) and three data- and parameter-efficient baseline models from literature \cite{Theisen2025DSC}: \justoliunet\ \cite{Justo2025SSS}, \mr\ \cite{Dempster2021MVF} and \hdcmr\ \cite{Schlegel2022HET}. All baseline models use a fully connected layer with softmax activation for classification. To maintain comparability we use the same classification module for \hsl-FC (see Sec. \ref{sec:models}). \hsl-CLF resembles the spectral classification model proposed together with \hsl\ \cite{Kong2025HSF}, using a more sophisticated classification module. 

We focus our training on \hyko\ and \hsidrive\ as the high-resolution of \gls*{hcv} images results in extreme training times for spectral classification. The results are presented in \tablename~\ref{tab:hs3-hyper-sc-comparison}. 

\begin{table}[h!]
    \centering
    \footnotesize
    \caption{Benchmark scores (\%) on the HS3-Bench \cite{Theisen2024HBS} test data for \gls{hsi}-only spectral classification models, \ie\ hyperspectral data as input and hyperspectral backbone models. Best results are bold, second-best are underlined.}
    \hspace*{-0.7cm}

    \begin{tabular}{lr|cccc}
                    \toprule
                       && \multicolumn{4}{c}{Testing}  \\\cmidrule(lr){3-6}
                       Dataset & Approach & \acrshort{oa} & \acrshort{aa} & $\fone$ & \acrshort{miou} \\
                                \midrule
                       \hyko & \justoliunet & 66.32 & 33.25 & 31.68 & 25.17 \\
                       \hyko & \mr & 69.99 & 40.47 & 40.77 & 31.74\\
                       \hyko & \hdcmr & 70.27 & 37.55 & 37.74 & 32.24 \\
                       \hyko & \hsl-FC & \underline{72.83} & \underline{43.78} & \underline{44.75} & \underline{34.79} \\
                       \hyko & \hsl-CLF & \textbf{79.56} & \textbf{54.30} & \textbf{55.26} & \textbf{44.63} \\
                       \midrule
                       \hsidrive & \justoliunet & \underline{78.81} & 26.15 & 26.29 & 21.38\\
                       \hsidrive & \mr & 75.14 & 31.17 & 31.34 & 24.11\\
                       \hsidrive & \hdcmr & 78.39 & \underline{31.49} & \underline{32.22} & \underline{25.20}\\
                       \hsidrive & \hsl-FC & 75.56 & 30.77 &  31.17 & 23.66\\   
                       \hsidrive & \hsl-CLF & \textbf{82.83} & \textbf{38.33} &  \textbf{39.70} & \textbf{31.87}\\   

                       \bottomrule
    \end{tabular}

    \label{tab:hs3-hyper-sc-comparison}
\end{table}

\hsl-CLF consistently achieves the best performance across all datasets and metrics, with improvements of up to 12\% \gls*{miou} on \hyko\ and up to 6\% \gls*{miou} on \hsidrive\ over the best baseline model. For \hyko\ the second best model is \hsl-FC, further proving the effectiveness of domain-transfer. On \hsidrive , \hdcmr\ and \mr\ slightly outperform \hsl-FC. When comparing the spectral classification results to semantic segmentation results in \tablename~\ref{tab:hs3-hyper-ss-comparison}, we observed a more significant performance drop on \hsidrive\  than on \hyko , indicating that spectral information is less discriminative fo \hsidrive , likely due to higher noise levels in near-infrared data. This makes spectral features extracted by \hsl\ less useful, giving more efficient \mr-based models an advantage. However, with better classification modules like the one used in \hsl-CLF, this advantage diminishes.

In summary, cross-domain knowledge transfer proved to be a highly effective strategy for \gls*{hsi} semantic segmentation and spectral classification, consistently improving performances over models relying solely on in-domain-in-modality training. 
However, models build on top of \gls*{hsi} foundation models, must be sufficiently robust to exploit the discriminative information encoded in their backbone features. While, \hsl-CLF \cite{Kong2025HSF} is the best \gls*{hsi}-only model for spectral classification, our proposed \hslrunet\ defines the state-of-the-art for \gls*{hsi}-only semantic segmentation. 

\subsection{Cross-Domain vs. Cross-Modality Transfer} \label{subsec:rgb-models}

To evaluate cross-domain and cross-modality approaches, we compared \hslrunet\ to models using \gls*{prgb}-projected \gls*{hsi} data (see \cite{Theisen2024HBS}) or RGB-based backbones \cite{Hurtado2025HAS} (\ie\ \dino\ \cite{Oquab2024DLR} and ImageNet-1K \cite{Deng2009IAL}).\tablename~\ref{tab:hs3-prgb-ss} shows the results, including reported results for \hsiadapter\ \cite{Hurtado2025HAS}, as the model was not yet publicly available when the experiments were performed.

\begin{table}[h!]
    \centering
    \caption{
    Benchmark scores (\%) on the HS3-Bench test data for models that are trained on 
    \gls*{prgb} projections or use RGB-pretrained backbones (\hsiadapter). Our \gls{hsi}-only cross-domain model, \hslrunet\ is listed for comparison. Best results per dataset are bold, second best are underscored.}
    \hspace*{-0.7cm}

    \begin{tabular}{llcr|cccc}
                    \toprule
                        
                       &&&& \multicolumn{4}{c}{Testing}  \\\cmidrule(lr){5-8}
                       Dataset & Approach & Backbone & Data & \acrshort{oa} & \acrshort{aa} & $\fone$ & \acrshort{miou}
                       \\\midrule
                       \gls*{hcv}  & \runet & -- & \gls*{prgb} & 87.95 & 56.65 & \underline{55.46} & 44.03 \\
                                 & \gls*{dl3} & -- & \gls*{prgb} & 87.00 & 55.33 & 54.08 & 42.58\\
                                 & \gls*{dl3} & ImageNet-1K & \gls*{prgb} & \underline{90.26} & \underline{64.10} & \textbf{61.93} & \underline{50.04}\\
                                 & \hsiadapter\  & \dino\ & \gls*{hsi} & \textbf{91.54} & \textbf{72.48} & -- & \textbf{58.81} \\
                                 & \runet\ & \hsl  & \gls*{hsi} & 88.17 & 57.53 & 55.12 & 43.92 \\
                                 \midrule
                       \hyko & \runet & -- & \gls*{prgb} & 89.18 & 73.92 & \underline{75.04} & 64.67\\
                                 & \gls*{dl3} & -- & \gls*{prgb} & 84.64 & 65.30 & 66.56 & 54.82\\      
                                 & \gls*{dl3} & ImageNet-1K & \gls*{prgb} & \underline{90.49} & \underline{74.87} & \textbf{77.11} & \underline{66.77}\\
                                 & \hsiadapter\ & \dino\ & \gls*{hsi} & \textbf{93.15} & \textbf{84.79} & -- & \textbf{77.14} \\
                                 & \runet\  & \hsl  & \gls*{hsi} & 88.24 &  71.21 & 71.78 & 61.54 \\
                                \midrule
                       \hsidrive & \runet & -- & \gls*{prgb} & 96.32 & 82.70 & 84.91 & 75.31 \\
                                 & \gls*{dl3} & -- & \gls*{prgb} & 92.74 & 66.59 & 69.46 & 57.84\\
                                 & \gls*{dl3} & ImageNet-1K & \gls*{prgb} & \underline{97.09} & 83.93 & \textbf{86.41} & \underline{77.44}\\
                                 & \hsiadapter\ & \dino\ & \gls*{hsi} & \textbf{99.27} & \textbf{96.47} & -- & \textbf{93.80} \\
                                 & \runet\ & \hsl & \gls*{hsi} & 96.78 & \underline{84.14} & \underline{85.60} & 76.33  \\
                                 \midrule \midrule
                            \textbf{Average} & \runet & -- & \gls*{prgb} & 91.15 & 71.09 & \underline{71.80} & 61.34 \\
                            \textbf{Perf.} & \gls*{dl3} & -- & \gls*{prgb} & 88.13 & 62.41 & 63.37 & 51.75 \\
                                    & \gls*{dl3} & ImageNet-1K & \gls*{prgb} & \underline{92.61} & \underline{74.30} & \textbf{75.15} & \underline{64.75}\\
                                    & \hsiadapter\  & \dino\ & \gls*{hsi} & \textbf{94.65} & \textbf{84.58} & -- & \textbf{76.58} \\
                                    & \runet\  & \hsl & \gls*{hsi} & 91.06 & 70.89 & 70.83 & 60.60 \\

                       \midrule
                            \textbf{Worst-Case} & \runet & -- & \gls*{prgb} & 87.95 & 56.65 & \underline{55.46} & 44.03 \\
                            \textbf{Perf.} & \gls*{dl3} & -- & \gls*{prgb} & 84.64 & 55.33 & 54.08 & 42.58 \\
                                    & \gls*{dl3} & ImageNet-1K & \gls*{prgb} & \underline{90.26} & \underline{64.10} & \textbf{61.93} & \underline{50.04}\\
                                    & \hsiadapter & \dino\ & \gls*{hsi} & \textbf{91.54} & \textbf{72.48} & -- & \textbf{58.81} \\
                                    & \runet\  & \hsl  & \gls*{hsi} & 88.17 & 57.53 & 55.12 & 43.92 \\
                       \midrule
                       \bottomrule
    \end{tabular}
    \label{tab:hs3-prgb-ss}
\end{table}

\hsiadapter\ achieves the best performance, followed by \gls*{dl3} with an Image- Net-1K \cite{Deng2009IAL} backbone. While \gls*{dl3} uses \gls*{prgb}-projected input, \hsiadapter\ exploits full spectral information and combines it with knowledge from a \dino\ \cite{Oquab2024DLR} vision foundation model. \hslrunet\ remained competitive with \gls*{dl3} on \hsidrive, the only dataset capturing only near-infrared light, where RGB-backbones may be less effective. \hslrunet\ also achieved comparable performance to \gls*{prgb}-based models without vision foundation models as backbones. 

The continued advantage of cross-modality over \gls*{hsi}-only approaches -- even with cross-domain knowledge transfer -- presumably stems from their pretraining on vast datasets, the discriminability of human-made objects in RGB, inherent in driving scenes, and the lower dimensionality of RGB data compared to HSI. 

In summary, cross-domain knowledge transfer strongly improved results of \gls*{hsi}-only models and further narrowed the performance gap to cross-modality approaches, especially in spectral ranges outside the visual spectrum.

\subsection{Robustness to Limited Training Data} \label{subsec:data-efficiency}

\begin{figure}[h]
    \centering
    \includegraphics[width=0.32\linewidth]{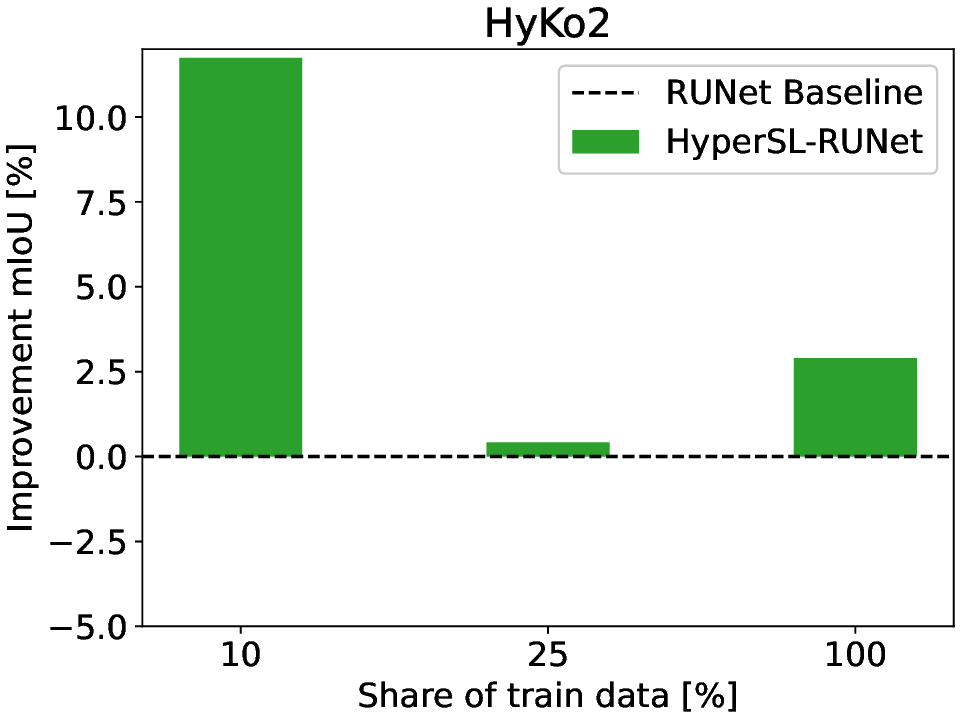}
    \includegraphics[width=0.32\linewidth]{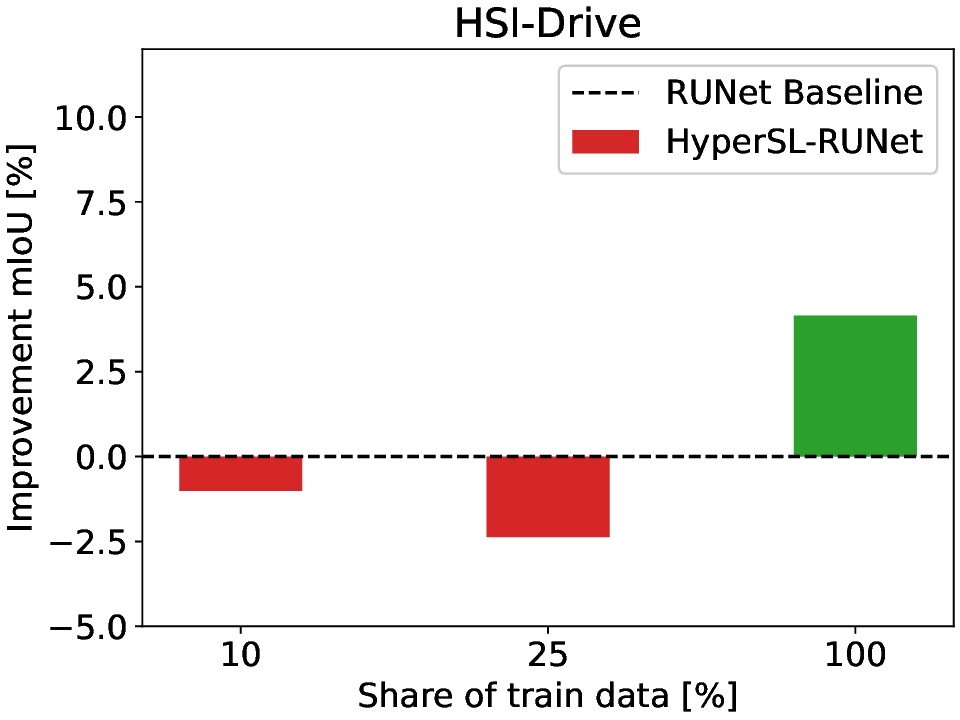}
    \includegraphics[width=0.32\linewidth]{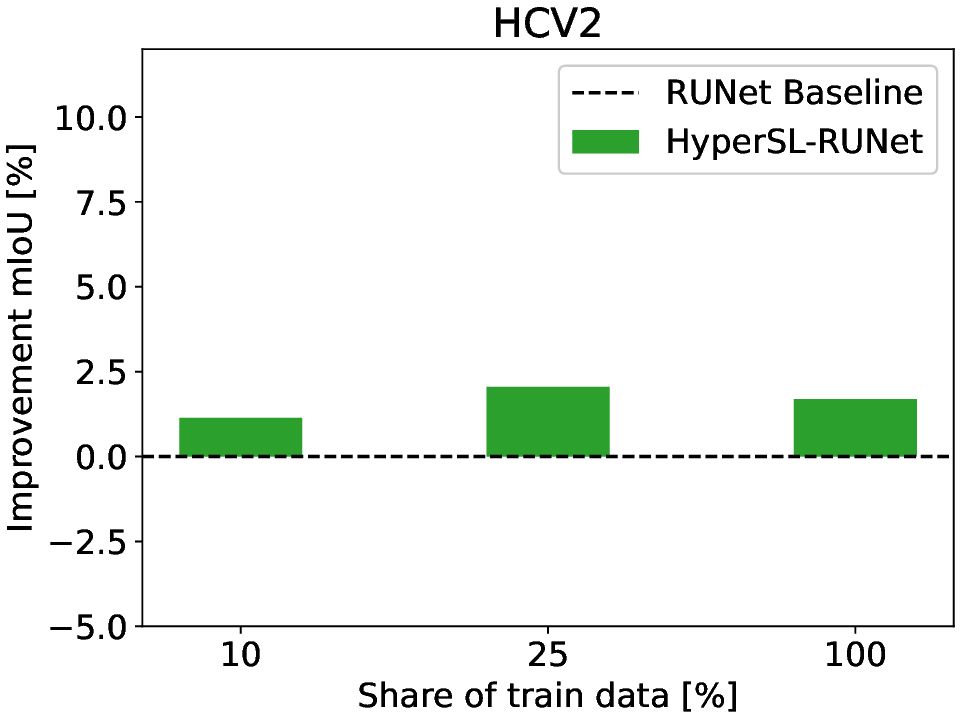}
    \caption{Improvements in mIoU-score for \hslrunet\ over \runet . Green bars indicate positive effect, red bar indicates negative effect.}
    \label{fig:delta-limited-data}
\end{figure}

To assess \hslrunet 's data efficiency, we compared models trained on fixed random subsets with 10\%, 25\% and 100\% the samples of the original training data. \figurename~\ref{fig:delta-limited-data} shows that \hslrunet\ consistently outperforms \runet\ for two out of three datasets in limited-data scenarios. 

An exception is \hsidrive , where \runet\ gives better results than \hslrunet\ for 10\% and 25\% subsets. During training we observed that \hslrunet 's validation scores converged faster initially, but \runet\ closed the gap over time. Given that our results in Sec. \ref{subsec:domain-transfer-ss} show that spectral information alone is less discriminative for \hsidrive, the spectral \hsl\ features were less useful and the simpler \runet\ could exploit spatial features more effectively given sufficient training time. However, \hslrunet 's better results for 100\% data on \hsidrive\ show that, with sufficient data, \hsl\ features improve performance.

\hslrunet\ achieved the biggest improvement in \gls*{miou} (+12\%) on \hyko\ at 10\% training data, where \runet\ failed to stabilize during training. While performance improvements for \hyko\ varied with training split size, for \gls*{hcv} they remained almost consistent (ca. +1.6\%). We also observed that the validation curves for \hslrunet\ were generally steeper and converged faster than those of \runet\ for all datasets and dataset sizes. The consistent and stable improvements observed for \gls*{hcv} could be attributed the high number of spectral channels and broad spectral coverage provides, which provide discriminative spectral information, that can be captured in \hsl\ features. Additionally, the characteristics of \gls*{hcv} are similar to those of remote sensing data used to train \hsl , explaining the observed model stability.

In summary, our observations highlight \hsl 's value in extracting discriminative features that help stabilize training and improve performance, even in limited-data settings. However, if spectral information plays only a secondary role (\eg\ due to noise in the spectra), with sufficient training time, the simpler \runet\ can be more effective for low-data scenarios.
\section{Conclusion} \label{sec:conclusion}

This paper introduces a taxonomy to categorize \gls*{hsi} semantic segmentation approaches based on the origin of their encoded knowledge: in-domain-in-modality training, cross-modality knowledge transfer and cross-domain knowledge transfer approaches. This taxonomy reveals cross-domain knowledge transfer as an underexplored yet promising strategy for \gls*{hsi} semantic segmentation. We systematically evaluate and compare all three strategies using the HS3-Bench benchmark, demonstrating the effectiveness of cross-domain knowledge transfer.

Our experiments reveal that cross-domain knowledge transfer is highly effective for both, semantic segmentation and spectral classification. Compared to models relying solely on in-domain-in-modality training, our approach consistently improves performance -- particularly in class-averaged metrics, suggesting greater robustness for minority classes. \hslrunet\ achieves an average 3\% improvement in \gls*{miou} over baseline models, with even larger gains on individual datasets. In spectral classification, where spatial context is absent, performance improvements of up to 10\% \gls*{miou} over baseline models were observed when paired with an appropriate classification head. This underscores the importance of spectral features in scenarios where spatial redundancy cannot compensate for prediction uncertainties.  

When comparing cross-modality and cross-domain approaches, \hslrunet\ remains competitive against models using \gls*{prgb}-projected \gls*{hsi} data and narrows the performance gap to cross-modality approaches. Cross-modality models based on vision foundation backbones still outperform \hslrunet\ in most cases, except for near-infrared dataset \hsidrive. This is likely due to vision models extensive pretraining and alignment with human-designed environments. Hence, \hslrunet 's ability to capture fine spectral details offers a complementary strength, making it especially valuable in applications where material discrimination or non-visible spectral bands are critical.  

The robustness of cross-domain knowledge transfer was further validated in limited-data scenarios, where \hslrunet\ outperformed baseline models trained on as little as 10\% of the original training data. Additionally, \hsl 's spectral tokenizer enables flexible applications across sensors with varying number of channels and spectral ranges without retraining, making it particularly appealing for real-world deployment, where acquiring large labeled datasets can be challenging. 

For future work, several directions seem promising. First, integrating \gls*{hsi} and vision foundation models into a single architecture could leverage the strength of both approaches, further enhancing performance. Second, following \cite{Laprade2025GPS}, compiling sufficient data to develop \gls*{hsi} foundation models tailored for proximal sensing could lead to adoption in unexplored domains. Finally, extending \hsl 's application to further domains, such as agriculture, environmental monitoring or recycling, could reveal a broader applicability beyond driving scenes. 

In summary, we demonstrate that cross-domain knowledge transfer improves performance over in-domain-in-modality training and narrows the performance gap to cross-modality approaches. Our proposed cross-domain \gls*{hsi} semantic segmentation model, \hslrunet , achieves state-of-the-art results among \gls*{hsi}-only methods and matches the performance of cross-modality approaches that do not rely on vision foundation models. Though cross-modality approaches still dominate, cross-domain knowledge transfer is particularly advantageous when discrimination relies on narrow spectral bands or non-visual spectral information. This work represents a step toward more practical, adaptable hyperspectral analysis in real-world settings, making \gls*{hsi} applications more accessible and effective.
\subsubsection*{Acknowledgements} This work was partially funded by Wehrtechnische Dienststelle 41 (WTD), Koblenz, Germany.
\printbibliography
\end{document}